\documentclass[letterpaper, 10 pt, conference]{ieeeconf}  

\IEEEoverridecommandlockouts                              

\usepackage{graphicx} 
\usepackage{amsmath} 
\usepackage{amsfonts,amssymb} 
\usepackage{float}
\usepackage{makecell}
\usepackage{setspace}
\usepackage{cite}
\usepackage[ruled,linesnumbered]{algorithm2e}
\usepackage{amsmath}
\usepackage{caption}
\usepackage{svg}
\usepackage{pdfpages}
\makeatletter
\let\NAT@parse\undefined
\makeatother
\usepackage[
colorlinks=true,
linkcolor=black,      
anchorcolor=black,  
citecolor=black, 
urlcolor = black,       
]{hyperref}

\title{\LARGE \bf
Highly dynamic locomotion control of biped robot enhanced by swing arms
}

\author{Weijie Wang\textsuperscript{1}, Song Liu\textsuperscript{1}, Qinfeng Shan\textsuperscript{2} and Lihao Jia\textsuperscript{*}
\thanks{W.Wang, Song Liu, Q.Shan, and L.Jia are with the Research Center for Brain-inspired Intelligence, Institute of Automation, Chinese Academy of Sciences, Beijing 100190 China and Centre for Artificial Intelligence and Robotics, Hong Kong Institute of Science \& Innovation, Chinese Academy of Sciences. Q.Shan and W.Wang are also with the School of Artificial Intelligence, University of Chinese Academy of Sciences, Beijing 100049 China.
	{\tt\small e-mail:wangweijie2020@ia.ac.cn.}}
\thanks{*Corresponding author is Lihao Jia.}%
}

\begin{document}

\maketitle
\thispagestyle{empty}
\pagestyle{empty}

\begin{abstract}
Swing arms have an irreplaceable role in promoting highly dynamic locomotion on bipedal robots by a larger angular momentum control space from the viewpoint of biomechanics. Few bipedal robots utilize swing arms and its redundancy characteristic of multiple degrees of freedom due to the lack of appropriate locomotion control strategies to perfectly integrate modeling and control. This paper presents a kind of control strategy by modeling the bipedal robot as a flywheel-spring loaded inverted pendulum (F-SLIP) to extract characteristics of swing arms and using the whole-body controller (WBC) to achieve these characteristics, and also proposes a evaluation system including three aspects of agility defined by us, stability and energy consumption for the highly dynamic locomotion of bipedal robots. We design several sets of simulation experiments and analyze the effects of swing arms according to the evaluation system during the jumping motion of Purple (Purple energy rises in the east)V1.0, a kind of bipedal robot designed to test high explosive locomotion. Results show that Purple’s agility is increased by more than 10\%, stabilization time is reduced by a factor of two, and energy consumption is reduced by more than 20\% after introducing swing arms.

\end{abstract}

\section{INTRODUCTION}
Bipedal robot are more adaptable to complex terrain compared than the traditional wheeled and tracked robots because of their discrete ground support characteristic. Jumping motion is the basis of realizing various highly dynamic motions of bipedal robots. Although there is a long standing interest in the jumping motion of bipedal robots in terms of theory and practice, it is  still a challenging problem to achieve an agile, stable and efficient jumping motion.

In the 1980s, a kind of jumping motion control strategy for legged robots was proposed by Raibert, this control strategy decomposed the jumping motion into three components: hopping, moving forward and body attitude rotation, and each of these three components was controlled individually\cite{raibert1986legged},\cite{raibert1984hopping}. This control strategy plays a seminal role in the development of legged robots. A large number of advanced control strategies for legged robots have emerged in recent decades, a typical control strategy is to use the simplified model of spring loaded inverted pendulum (SLIP)\cite{schwind1998spring} to plan the trajectory of bipedal robot's center of mass (COM) and then design the corresponding joint controller to track the COM trajectory\cite{tamaddoni2010biped},\cite{cho2011online}. Some other control strategies directly use the momentum principle to plan the joint trajectory and then design the corresponding joint controller to track the joint trajectory\cite{kajita2004hop}\cite{otani2018trunk}.

Although these control strategies are effective, none of them take swing arms into account. Many biomechanical studies have shown that swing arms do improve the performance of jumping motion: jumping height\cite{hara2006effect}\cite{lees2004understanding}\cite{ashby2006optimal}, landing stability\cite{pijnappels2010armed}\cite{meyns2013and}, energy consumption\cite{meyns2013and}\cite{kaddar2015arm}\cite{collins2009dynamic}\cite{park2008synthesis}\cite{de2019influence}. There also exist many research teams that have studied the effects of swing arms in bipedal robot motion, however, most of them analyzed the effects of swing arms of bipedal robots, such as posture stability and movement speed, based on the non-highly dynamic motion\cite{sobajima2013bipedal}\cite{zhang1effectiveness}\cite{miyata2019walking}\cite{park2021whole}\cite{yang2014yaw}. Although a few research teams analyzed the role of swing arms based on the highly dynamic motion of bipedal robots, swing arms are not adequately considered in the modeling, the trajectory of swing arms is artificially designed\cite{mineshita2020jumping}. In summary, there is not yet a systematic and comprehensive analysis of the effects of swing arms during the highly dynamic motion of bipedal robots.

In respect to the shortcomings of the above studies, this paper proposes a kind of control strategy which can fully utilize swing arms for the highly dynamic motion of bipedal robots and defines a evaluation system which takes three aspects of agility defined by us, stability and energy consumption into account for the highly dynamic motion of bipedal robots. The control strategy we proposed integrates swing arms into the whole realization process of the jumping motion of bipedal robots: modeling, trajectory planning and trajectory tracking. In the modeling. We equated swing arms to a flywheel and obtained F-SLIP model\cite{pratt2006capture} by adding the flywheel to SLIP model, which can capture characteristics of swing arms. In the trajectory planning, we chose F-SLIP as the model to plan the jumping motion trajectory for taking into account the effects of swing arms. In the trajectory tracking, we selected the WBC to track the jumping motion trajectory, WBC is a kind of control method  based on the optimization and full model of robot. So it can intelligently generate proper joint control inputs according to different control tasks, which can take full advantage of characteristics of swing arms.\cite{lees2004understanding}.

The remainder is organized as follows. The Section II describes the F-SLIP model and dynamics model of bipedal robot Purple V1.0. The Section III plans the desired jumping motion trajectory based on the F-SLIP model. the Section IV designs the WBC to track the jumping motion trajectory. The Section V presents the simulation results and analyses. The Section VI gives conclusions and future research work.
\section{Modeling}
In this section, We first present the F-SLIP Model for planning the jumping motion trajectory. Then we model the dynamics model of the bipedal robot for constructing the WBC controllers.
\subsection{The F-SLIP Model}
\begin{figure}[htb]
	\begin{center}
		\includegraphics[width=2 in,trim=30 0 0 0]{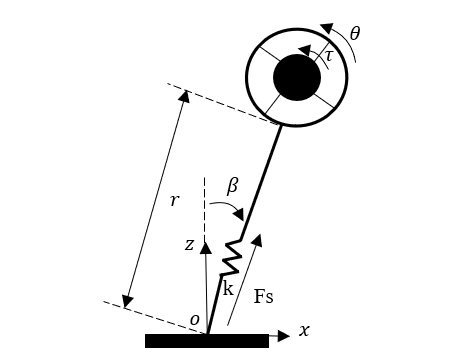}
	\end{center}
	\caption{Fig. 1. The F-SLIP model.}\label{F-SLIP}
\end{figure}

The F-SLIP is the SLIP model\cite{schwind1998spring} with a flywheel attached on the mass, as shown in Fig. \ref{F-SLIP}. The part of the F-SLIP model's point mass dynamics captures the hopping dynamics of COM, the flywheel represents the rotational part of the centroidal dynamics of bipedal robot. The dynamics are shown below:
\begin{align}
m \ddot{x} &=-\frac{\tau}{r} \cos (\beta)+F_{s} \sin (\beta) \\
m \ddot{z} &=-mg+\frac{\tau}{r} \sin (\beta)+F_{s} \cos (\beta) \\
I \ddot{\theta} &=\tau \\
F_{s}&=k\left(r_{0}-r\right)
\end{align}
Where:
\hspace*{1em}
\begin{itemize}
    \item $k$ is the stiffness of the spring.
    \item $m$ is the total mass.
    \item $I$ is the rotational inertia of the flywheel.
    \item $\tau$ is the torque on the flywheel.
    \item $\theta$ is the angle of the flywheel.
    \item $\beta$ is the angle of the leg.
    \item $r$ is the leg length.
    \item $r_0$ is the natural leg length. 
    \item $F_s$ is the spring force. 
    \item $x$ is the horizontal position of COM.
    \item $z$ is the vertical position of COM.
\end{itemize}

\subsection{The Robot Model}
\begin{figure}[htb]
	\begin{center}
		\includegraphics[width=1 in,trim=30 0 0 0]{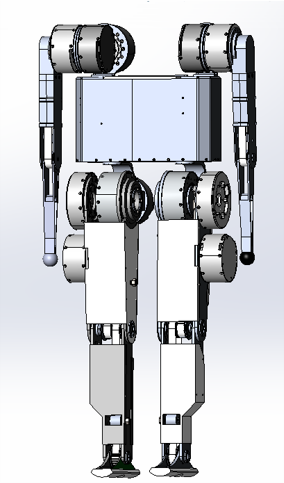}
	\end{center}
	\caption{Fig. 2. Purple V1.0.}
	\label{zidong}
\end{figure}

The bipedal robot we use is called Purple V1.0 as shown in Fig. \ref{zidong}, it has 22 degrees of freedom. Let $q\in SE(3) \times \mathbb{R}^{16}$ represent the configuration of the robot. The Euler-Lagrange equation of the robot's dynamics are shown below:
\begin{equation}
\begin{aligned}
&M(q) \ddot{q}+V(q, \dot{q})+G(q)=S_a^T\tau+J_{s, p}^{T} F_{s, p} \\
&J_{h,p}(q) \ddot{q} +\dot{J}_{h,p}(q)\dot{q}=0
\end{aligned}
\end{equation}
Where $M(q)$ is the mass matrix, $V(q, \dot{q})$ is the Coriolis, centrifugal term, $G(q)$ is the actuation matrix, $S_a$ is the actuation matrix, $\tau \in \mathbb{R}^{16}$ is the motor torque vector, $F_{s, p}$ is the foot contact force vector, $J_{s, p}$ is the corresponding Jacobian of the foot contact force, $p$ indicates different motion phases.
\section{Trajectory Planning}
In this section, we planned the jumping motion trajectory based on the F-SLIP model. In the jumping and landing phase, the COM and leg angle trajectories are the key parameters considered. We constructed the trajectory planning as optimization problem, it can be described and solved using the direct transcription method\cite{betts1998survey}, a trapezoidal integration is used to integrate the dynamics from the discrete trajectory. In the flight phase, we completed the landing angle calculation and then plan the trajectory of the leg angle.

\subsection{Jumping Phase Trajectory Planning}
The jumping phase is defined as the duration from the start of the jumping motion until the feet leave the ground. Achieving desired vertical and horizontal velocity is the main objective in the jumping phase. In addition, it is necessary to maintain the stability of the robot's posture. Therefore, we constructed constraints and cost functions as follows.
\subsubsection{Velocity Constraints} Since we constructed the jumping motion in the sagittal plane, so we only applied constraints on horizontal and vertical velocity as given by:
\begin{align}
\dot{x}^{tf}=\dot{x}^{d}\\
\dot{z}^{tf}=\dot{z}^{d}
\end{align}
Where $\dot{x}^{tf}$ and $\dot{z}^{tf}$ are horizontal and vertical velocity of COM at takeoff respectively, $\dot{x}^{d}$ and $\dot{z}^{d}$ are desired horizontal and vertical velocity of COM respectively.
\subsubsection{Swing Arms Constraints} The angular velocity and angular acceleration of the flywheel must be constrained for obtaining a proper arm-swing motion due to the limited output torque of swing arms. the swing arms constraints are give by: 
\begin{align}
\dot{\theta}_u > \dot{\theta} > \dot{\theta}_l\\
\ddot{\theta}_u > \ddot{\theta} > \ddot{\theta}_l
\end{align}
Where $\dot{\theta}_u$ and $\ddot{\theta}_u$ are the maximum angular velocity and acceleration of swing arms respectively, $\dot{\theta}_l$ and $\ddot{\theta}_l$ are the minimum angular velocity and acceleration of swing arms respectively.

\subsubsection{Stability Constraint} In order to keep the robot from slipping, we established the stability constraint as follows:
\begin{equation}
-\mu F_{z}  \leq F_{x} \leq \mu F_{z}
\end{equation}
where $F_{x}$ is the horizontal ground reaction force, $F_{z}$ is the vertical ground reaction force, and $\mu$ is the friction coefficient.
\subsubsection{Cost}In order to take full advantage of the energy-saving property of swing arms\cite{kaddar2015arm}, we set the optimization objective of the jumping phase to minimize the virtual energy, as follows:
\begin{equation}
J_{\text {Jumping }}=\sum_{i}\left(c_{1} \ddot{r}_{i}^{2}+c_{2} \tau_{i}^{2}\right) \Delta T \label{(12)}
\end{equation}
where the discretized variables are indexed by $i$, $\Delta T$ is the time discretization and $c_{1}$ and $c_{2}$ represent weightings of the cost.
\subsection{Flight Phase Trajectory Planning}
The flight phase of the jumping motion is defined as the duration from the time of the take-off to the time when the feet strike the ground. The goal of the flight phase is to adjust the attitude and achieve a proper foothold. The trajectory planning of the flight phase includes two parts: one is to calculate the proper landing angle, the other is to plan the flight trajectory of the leg angle.

we calculated the proper landing angle based on the concept of capture point proposed by Pratt et al in 2006\cite{pratt2006capture}, as shown below: 
\begin{align}
\beta^{Landing}&=cos^{-1}({x}_{cp}/r)\\
{x}_{cp}&=\dot x^{Landing} \sqrt{\frac{z^{Landing}}{g}}
\end{align}
where ${x}_{cp}$ is the desired horizontal position of COM with respect to the foot at the moment of landing, $\dot{x}^{Landing}$ is the horizontal velocity of COM at the moment of landing, $z^{Landing}$ is the vertical position of COM at the moment of landing, r is the leg length, $\beta^{Landing}$ is the landing angle.
Then we plan the flight trajectory of the leg angle, we keep the leg length constant and interpolate the landing angle to obtain the flight trajectory $\beta^{Flight}=[\beta_0^{Flight},...,\beta_N^{Flight}]$, $\beta_0^{Flight}$ is the leg angle at the moment of the take-off, $\beta_N^{Flight}=\beta^{Landing}$ is the landing angle.
\subsection{Landing Phase Trajectory Planning}
The landing phase of the jumping motion was defined as the duration from the time of landing to the time when the robot rests to a static configuration. In the landing phase, the stability constraints, swing arms constraints and cost are the same to that of the jumping phase. We just shown the static configuration constraints.
\subsubsection{Static Configuration Constraints} The robot remains stationary above the foot in the final stage of the landing phase, the constraints are established as follows:
\begin{align}
z_N^{Landing}&=z^{d}\\
\dot{z}_N^{Landing}&=x_N^{Landing}=\dot{x}_N^{Landing}=\dot{\theta}_N^{Landing}=0\\
{F_s}_N^{Landing}&=mg
\end{align}
where $N$ indicates landing phase's final state, $z^d$ is the desired vertical position of the robot in the final stage of the landing phase. 

\section{Trajectory tracking}
In this section, we constructed the WBC, which is described and solved by quadratic programming (QP), to track the jumping motion trajectory. Variables for the programming are joint accelerations $\ddot{q}$ , joint torques $\tau$ and contact force $F_{s,p}$.

\subsection{Jumping Phase WBC}
In order to achieve the desired performance of the jumping motion, the robot needs to be able to accurately track the COM trajectory including the COM position, COM velocity, and COM acceleration. Meanwhile, the rotational angular momentum of the robot needs to be controlled as small as possible to ensure a proper posture of take-off to prevent the robot from overturning. So we constructed two tasks: COM trajectory tracking task and  attitude rotation control task.
\subsubsection{COM trajectory tracking task} We charactered this task by the joint space\cite{khatib1987unified} and related the task velocity $\boldsymbol{\dot{p}}$, the velocity of COM, to the joint rates $\boldsymbol{\dot{q}}$ by the Jacobian dynamics:
\begin{align}
\boldsymbol{\dot{p}}=\boldsymbol{J}_{p} \boldsymbol{\dot{q}}
\end{align}
where $\boldsymbol{J}_{p}$ is the Jacobian matrix of COM velocity. However, in order to complete the COM trajectory tracking task, we needed to find joint torques that result in joint accelerations $\ddot{\boldsymbol{q}}$ with:
\begin{align}
{\boldsymbol{\ddot p}}=\boldsymbol{J}_{p} {\boldsymbol{\ddot q}}+\boldsymbol{\dot{J}}_{p} {\boldsymbol{\dot q}}
\end{align}
such that ${\boldsymbol{\ddot p}}$ most closely matches the commanded COM trajectory tracking task dynamics $\boldsymbol{\ddot{p}}_c$:
\begin{align}
\boldsymbol{\ddot{p}}_c=\boldsymbol{\ddot{p}}_{d}+\boldsymbol{K}_{D, p}\left({\boldsymbol{\dot p}}_{d}-{\boldsymbol{\dot p}}\right)+\boldsymbol{K}_{P, p}\left(\boldsymbol{p}_{d}-\boldsymbol{p}\right)
\end{align}
where $\boldsymbol{{p}}_{d}$, $\boldsymbol{\dot{p}}_{d}$ and $\boldsymbol{\ddot{p}}_{d}$ are the desired position, velocity and acceleration of COM respectively, $\boldsymbol{K}_{D, p}$ is the COM velocity gain matrix, $\boldsymbol{K}_{P, p}$ is the COM position gain matrix.

\subsubsection{Attitude rotation control task} In the same way as in the COM trajectory tracking task, the relation between the pitch component of the centroidal angular momentum and joint rates $\boldsymbol{\dot{q}}$ should be expressed as follows:
\begin{align}
{{h}}_r=\boldsymbol{A}_{r} \boldsymbol{\dot{q}}
\end{align}
where ${h}_r$ is the pitch component of the centroidal angular momentum, $\boldsymbol{A}_{r}$ is the pitch component of the centroidal momentum matrix\cite{orin2008centroidal}. We need to find joint torques that result in joint accelerations $\ddot{\boldsymbol{q}}$ with:
\begin{align}
{\dot{h}}_r=\boldsymbol{A}_{r} \boldsymbol{\ddot{q}}+\boldsymbol{\dot{{A}}}_{r} \boldsymbol{\dot{q}}
\end{align}
such that ${\dot{h}}_r$ must approach 0.
\subsubsection{Costs} In the jumping phase, cost functions are described as follows:
\begin{align}
&\| {\boldsymbol{\ddot p}}-\boldsymbol{\ddot{p}}_{c}\|_{\boldsymbol{W}_p}\label{jumping wbc cost1}\\ 
&\| {\dot{h}}_r\|_{\boldsymbol{W}_r}\label{jumping wbc cost2}
\end{align}

where $\boldsymbol{W}_p$ is the weight matrix of the COM trajectory tracking task, $\boldsymbol{W}_r$ is the weight matrix of the attitude rotation control task.
\subsection{Flight Phase WBC}
In the flight phase, we transformed the flight trajectory into the foot trajectory for the convenient description of task, and we controlled the robot track the foot trajectory. Apart from that, we must keep the foot level for landing steadily during the whole flight phase. Consequently, we constructed two tasks: the foot trajectory tracking task and foot level control task.
\subsubsection{Foot trajectory tracking task}
We charactered the foot trajectory tracking task by the joint space\cite{khatib1987unified} and related the task velocity $\boldsymbol{\dot{p}}_f$, the velocity of foot, to joint rates $\boldsymbol{\dot{q}}$ by the Jacobian  dynamics :
\begin{align}
\boldsymbol{\dot{p}}_f=\boldsymbol{J}_f \boldsymbol{\dot{q}}
\end{align}
where $\boldsymbol{J}_{f}$ is the Jacobian matrix of the foot velocity. In order to complete foot trajectory tracking task, we needed to find joint torques that result in joint accelerations $\ddot{\boldsymbol{q}}$ with:
\begin{align}
{\boldsymbol{\ddot p}}_f=\boldsymbol{ J}_{f} {\boldsymbol{\ddot q}}+\boldsymbol{\dot{J}}_{f} {\boldsymbol{\dot q}}
\end{align}
such that ${\boldsymbol{\ddot p}}_f$ most closely matches the commanded foot trajectory tracking task dynamics $\boldsymbol{\ddot{p}}_{fc}$:
\begin{align}
\boldsymbol{\ddot{p}}_{fc}=\boldsymbol{\ddot{p}}_{fd}+\boldsymbol{K}_{D, f}\left({\boldsymbol{\dot p}}_{fd}-{\boldsymbol{\dot p}}_f\right)+\boldsymbol{K}_{P, f}\left(\boldsymbol{p}_{fd}-\boldsymbol{p}_f\right)
\end{align}
where $\boldsymbol{{p}}_{fd}$, $\boldsymbol{\dot{p}}_{fd}$ and $\boldsymbol{\ddot{p}}_{fd}$ are the desired position, velocity and acceleration of foot respectively, $\boldsymbol{K}_{D, f}$ is the foot velocity gain matrix, $\boldsymbol{K}_{P, f}$ is the foot position gain matrix.

\subsubsection{Foot level control task} We represented the angular velocity of foot as a function of joint rates $\dot{q}$ : 
\begin{align}
\boldsymbol{\dot{\theta}}_f=\boldsymbol{J}_{f\omega} \boldsymbol{\dot{q}}
\end{align}
where $\boldsymbol{\dot{\theta}}_f$ is the angular velocity of foot, $\boldsymbol{J}_{f\omega}$ is Jacobian matrix of foot angular velocity. Joint torques should be calculated to find the joint acceleration $\ddot{q}$ with:
\begin{align}
{\boldsymbol{\ddot \theta}}_f=\boldsymbol{J}_{f\omega} {\boldsymbol{\ddot q}}+\boldsymbol{\dot{J}}_{f\omega} {\boldsymbol{\dot q}}
\end{align}
such that $\boldsymbol{\ddot{\theta}}_{f}$ most closely matches the commanded foot level task dynamics $\boldsymbol{\ddot{\theta}}_{fc}$:
\begin{align}
\boldsymbol{\ddot{\theta}}_{fc}=\boldsymbol{\ddot{\theta}}_{fd}+\boldsymbol{K}_{D, f\omega}\left({\boldsymbol{\dot {\theta}}}_{fd}-{\boldsymbol{\dot \theta}}_f\right)+\boldsymbol{K}_{P, f\omega}\left(\boldsymbol{\theta}_{fd}-\boldsymbol{\theta}_f\right)
\end{align}
where $\boldsymbol{{\theta}}_{fd}$, $\boldsymbol{\dot {\theta}}_{fd}$ and $\boldsymbol{\ddot{\theta}}_{fd}$ are the desired posture, angular velocity and angular accelerator of foot. $\boldsymbol{K}_{D, f\omega}$ is the foot angular velocity gain matrix,  $\boldsymbol{K}_{P, f\omega}$ is the foot posture gain matrix.
\subsubsection{Costs} In the flight phase, cost functions are described as follows:
\begin{align}
&\| {\boldsymbol{\ddot p}}_f-\boldsymbol{\ddot{p}}_{fc}\|_{\boldsymbol{W}_{f}}\label{flight wbc cost1}\\
&\| {\boldsymbol{\ddot \theta}_f}-\boldsymbol{\ddot{\theta}}_{fc}\|_{\boldsymbol{W}_{f\omega}}&\label{flight wbc cost2}
\end{align}
where $\boldsymbol{W}_{f}$ is the weight matrix for the foot trajectory tracking task, $\boldsymbol{W}_r$ is the weight matrix for the foot level task.

\subsection{Landing Phase WBC}
In order to land softly and smoothly, bipedal robot must be able to keep the torso level and track the COM reference trajectory, we constructed two tasks: the COM trajectory tracking task as described in the section jumping phase WBC and torso level control task. 
\subsubsection{Torso level control task} We represented the angular velocity of the robot torso as follows: 
\begin{align}
\boldsymbol{\dot{\theta}}_t=\boldsymbol{J}_{t\omega} \boldsymbol{\dot{q}}
\end{align}
where $\boldsymbol{\dot{\theta}}_t$ is the angular velocity of the torso, $\boldsymbol{J}_{t\omega}$ is the Jacobian matrix of the torso angular velocity. Joint torques should be calculated to find joint accelerations $\ddot{q}$ with:
\begin{align}
{\boldsymbol{\ddot \theta}}_t=\boldsymbol{J}_{t\omega} {\boldsymbol{\ddot q}}+\boldsymbol{\dot{J}}_{t\omega} {\boldsymbol{\dot q}}
\end{align}
such that $\boldsymbol{\ddot{\theta}}_{t}$ most closely matches the commanded torso level control task dynamics $\boldsymbol{\ddot{\theta}}_{tc}$:
\begin{align}
\boldsymbol{\ddot{\theta}}_{tc}=\boldsymbol{\ddot{\theta}}_{td}+\boldsymbol{K}_{D, t\omega}\left({\boldsymbol{\dot {\theta}}}_{td}-{\boldsymbol{\dot \theta}}_t\right)+\boldsymbol{K}_{P, t\omega}\left(\boldsymbol{\theta}_{td}-\boldsymbol{\theta}_t\right)
\end{align}
where $\boldsymbol{{\theta}}_{td}$, $\boldsymbol{\dot {\theta}}_{td}$ and $\boldsymbol{\ddot{\theta}}_{td}$ are the desired posture, angular velocity and angular accelerator of torso respectively, they should be set to 0. $\boldsymbol{K}_{D, t\omega}$ is the torso angular velocity gain matrix,  $\boldsymbol{K}_{P, t\omega}$ is the torso posture gain matrix.
\subsubsection{Costs} In the landing phase, cost functions are described as follows:
\begin{align}
&\| {\boldsymbol{\ddot p}}-\boldsymbol{\ddot{p}}_{c}\|_{\boldsymbol{W}_p} \label{COM tracking task in landing}\\
&\| {\boldsymbol{\ddot \theta}}_t-\boldsymbol{\ddot{\theta}}_{tc}\|_{\boldsymbol{W}_{t\omega}}
\end{align}
where $\boldsymbol{W}_{t\omega}$ is the weight matrix for the torso level control task, $\boldsymbol{\ddot{p}}$, $\boldsymbol{\ddot{p}}_c$ and $\boldsymbol{W_p}$ are the same of (\ref{jumping wbc cost1}).

\section{RESULTS AND DISCUSSION}
\begin{figure}[htb]
	\begin{center}
		\includegraphics[width=3 in,trim=30 0 0 0]{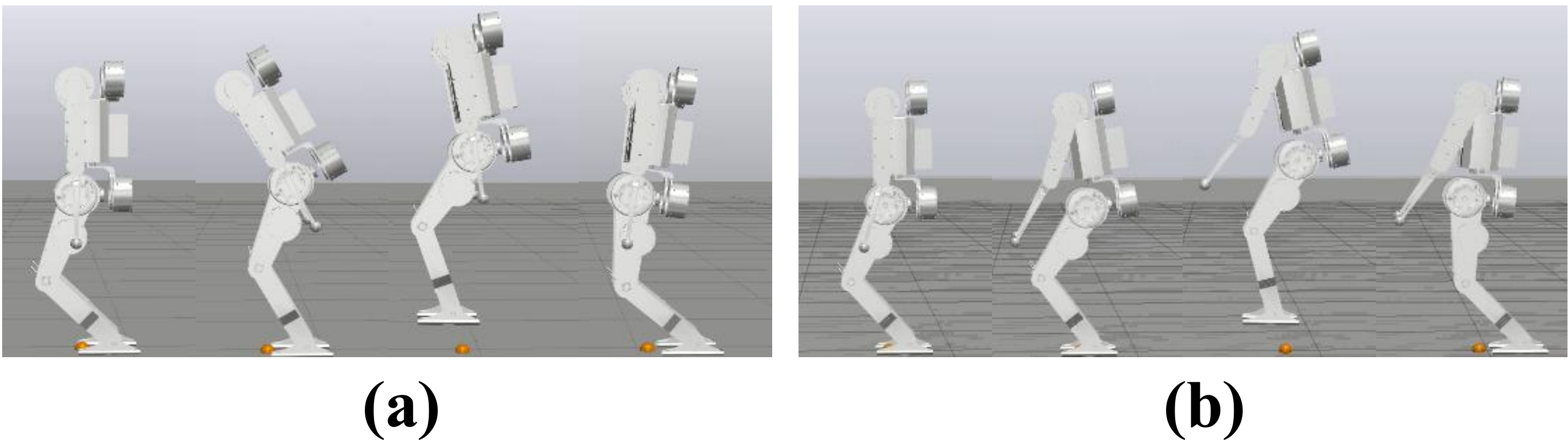}
	\end{center}
	\caption{Fig. 3. Images of simulations. (a) is the jumping motion of NAS, (b) is the jumping motion of AS.}
	\label{Images of simulation}
\end{figure}

We verified the effectiveness of the proposed control strategy through simulations performed on the bipedal robot Purple V1.0 ( Fig. \ref{zidong}), we completed the jumping motions (height = 0.15m, distance = 0.1m) of two scenarios: one without arm-swing (NAS), the other with arm-swing (AS), as shown in Fig. \ref{Images of simulation}.

\subsection{Agility results}
In order to measure the agility, we defined the agility indicator for the highly dynamic motion agility (HDMA) according to the vertical jumping agility proposed by Haldane et al.\cite{haldane2016robotic} as follows:
\begin{align}
    \text{HDMA} = \frac{\sqrt{h^2+d^2}}{t_{stance}+t_{flight}}
\end{align}
where $h$ is the motion height, $d$ is the motion distance, $t_{stance}$ is the total time from the onset of actuation to the end of stance, $t_{flight}$ is the total time between the time when the robot leaves the ground and the time when it touches the ground.

\begin{figure}[htbp]
  \begin{minipage}[t]{0.5\linewidth}
    \centering
    \includegraphics[scale=0.29]{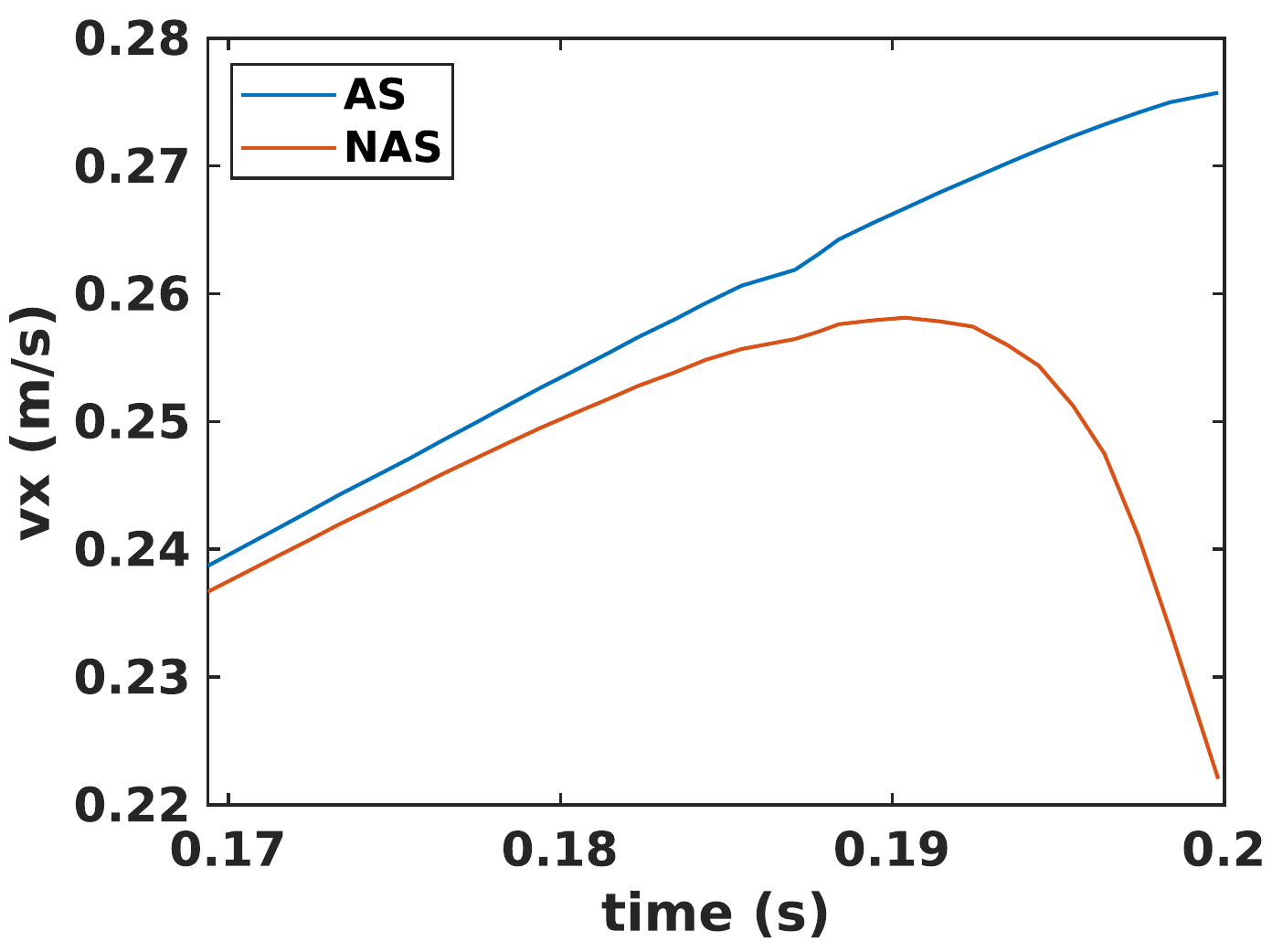}
    \caption{Fig. 4. vx in jumping phase.}
    \label{vx in jumping phase}
  \end{minipage}%
  \begin{minipage}[t]{0.5\linewidth}
    \centering
    \includegraphics[scale=0.29]{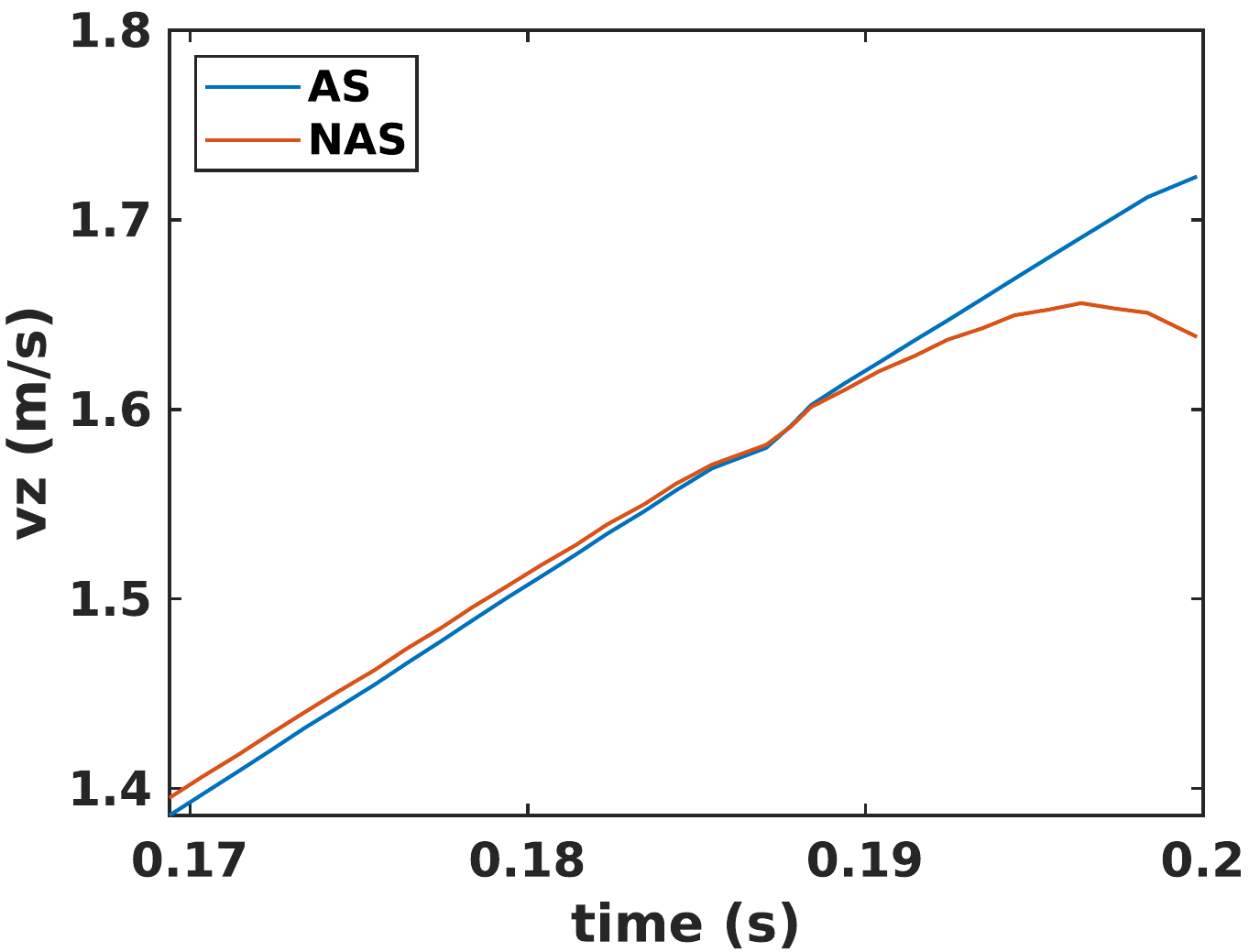}
    \caption{Fig. 5. vz in jumping phase.}
    \label{vz in jumping phase}
  \end{minipage}
\end{figure}

Then we calculated the HDMAs of both cases AS and NAS according to the COM velocity (Fig. \ref{vx in jumping phase}, Fig. \ref{vz in jumping phase}) as shown in TABLE \ref{HDMA}. We can see that the HDMA of the AS is approximate 10\% higher than that of the NAS.
\begin{table}[htbp]
	\caption{TABLE I. HDMA}
	\label{HDMA}
	\begin{center}
		\begin{tabular}{|c|c|}
			\hline
			AS & 0.29m/s\\ \hline
			NAS & 0.32m/s\\ \hline
		\end{tabular}
	\end{center}
\end{table}

\begin{figure}[htbp]
  \begin{minipage}[t]{0.5\linewidth}
    \centering
    \includegraphics[scale=0.3]{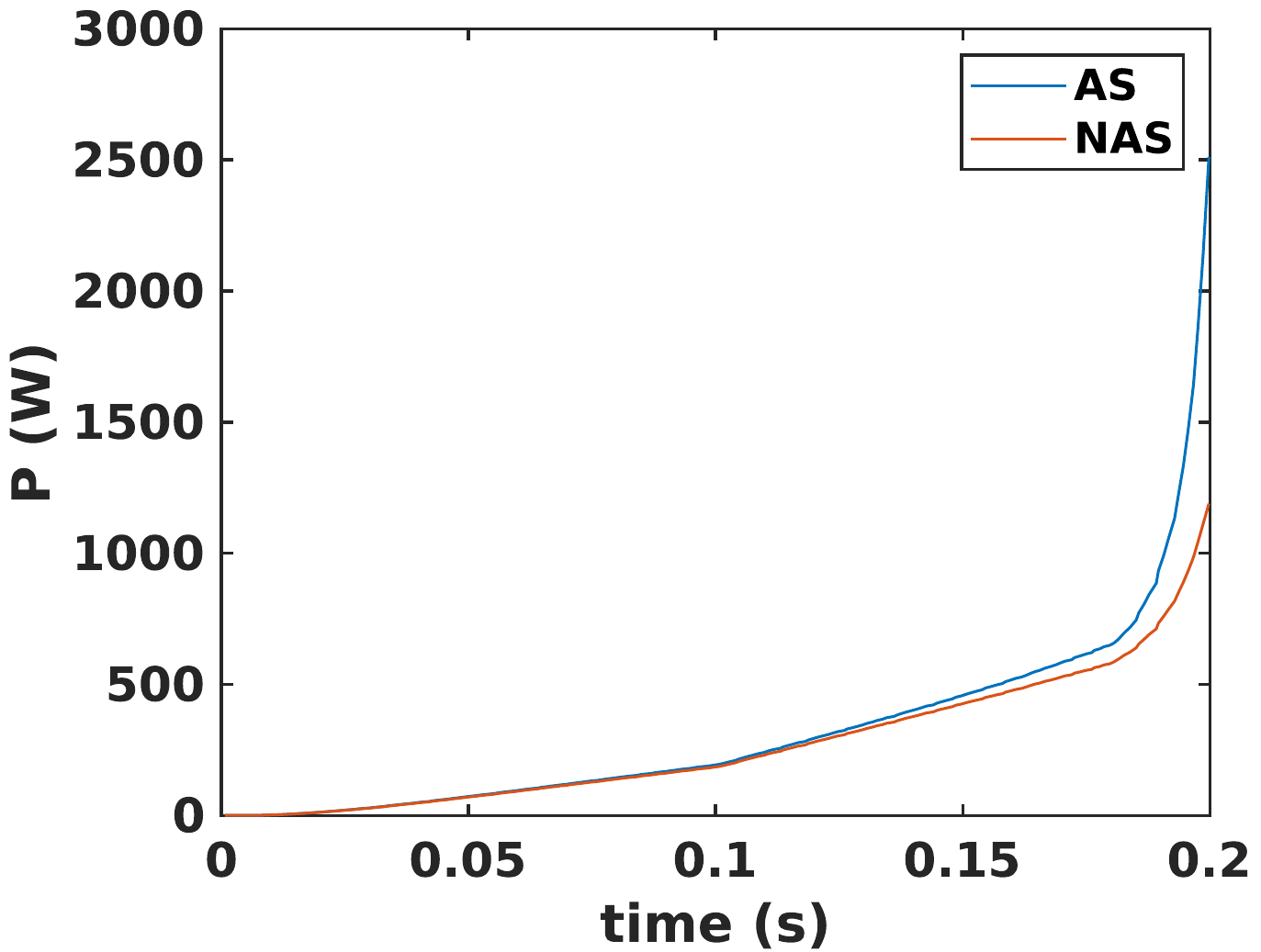}
    \caption{Fig. 6. Lower limbs power.}
    \label{Low limbs power}
  \end{minipage}%
  \begin{minipage}[t]{0.5\linewidth}
    \centering
    \includegraphics[scale=0.3]{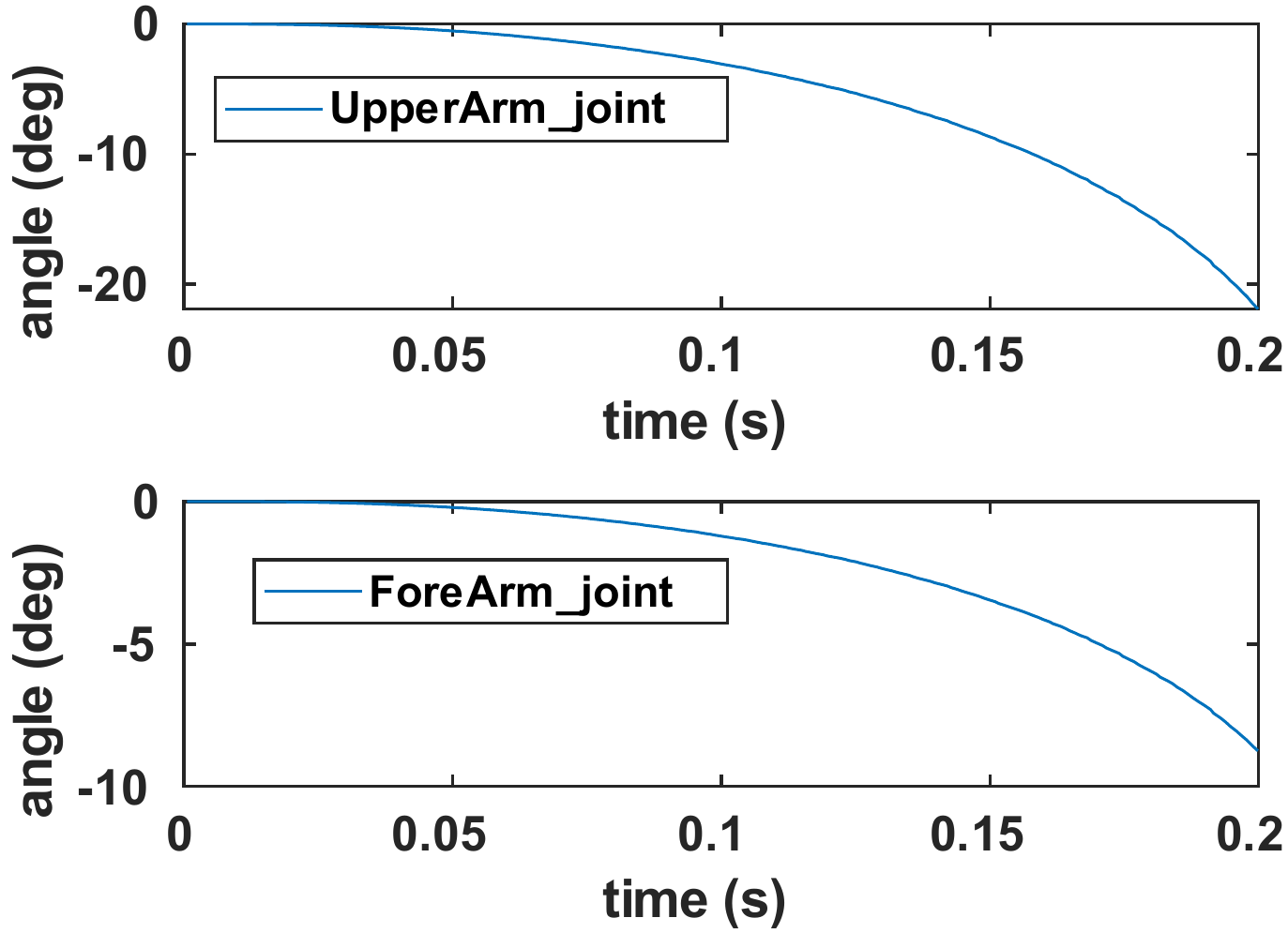}
    \caption{Fig. 7. Arm angles in jumping phase.}
    \label{Arms angle in stance phase}
  \end{minipage}
\end{figure}

Next, we obtained the lower limbs power of both cases AS and NAS (Fig. \ref{Low limbs power}) , joint angle of arms (Fig. \ref{Arms angle in stance phase}), It can be seen that the lower limbs power of the NAS has been higher than that of the AS throughout the jumping phase, but the COM velocity of the NAS is lower than that of the AS. In addition, we observed that arms have been swinging in one direction during the jumping phase, and the swing speed is getting more and more violent. In summary, the increase of agility is directly related to the arm-swing\cite{lees2004understanding},

\subsection{Stability results}
In order to measure the stability, we defined the stabilization time as the maximum time of vx, vz, pitch moment reaching static and the total landing steady-state error (TLSE) as the accumulation of steady-state error during the landing phase. 

\begin{figure}[htbp]
  \begin{minipage}[t]{0.5\linewidth}
    \centering
    \includegraphics[scale=0.29]{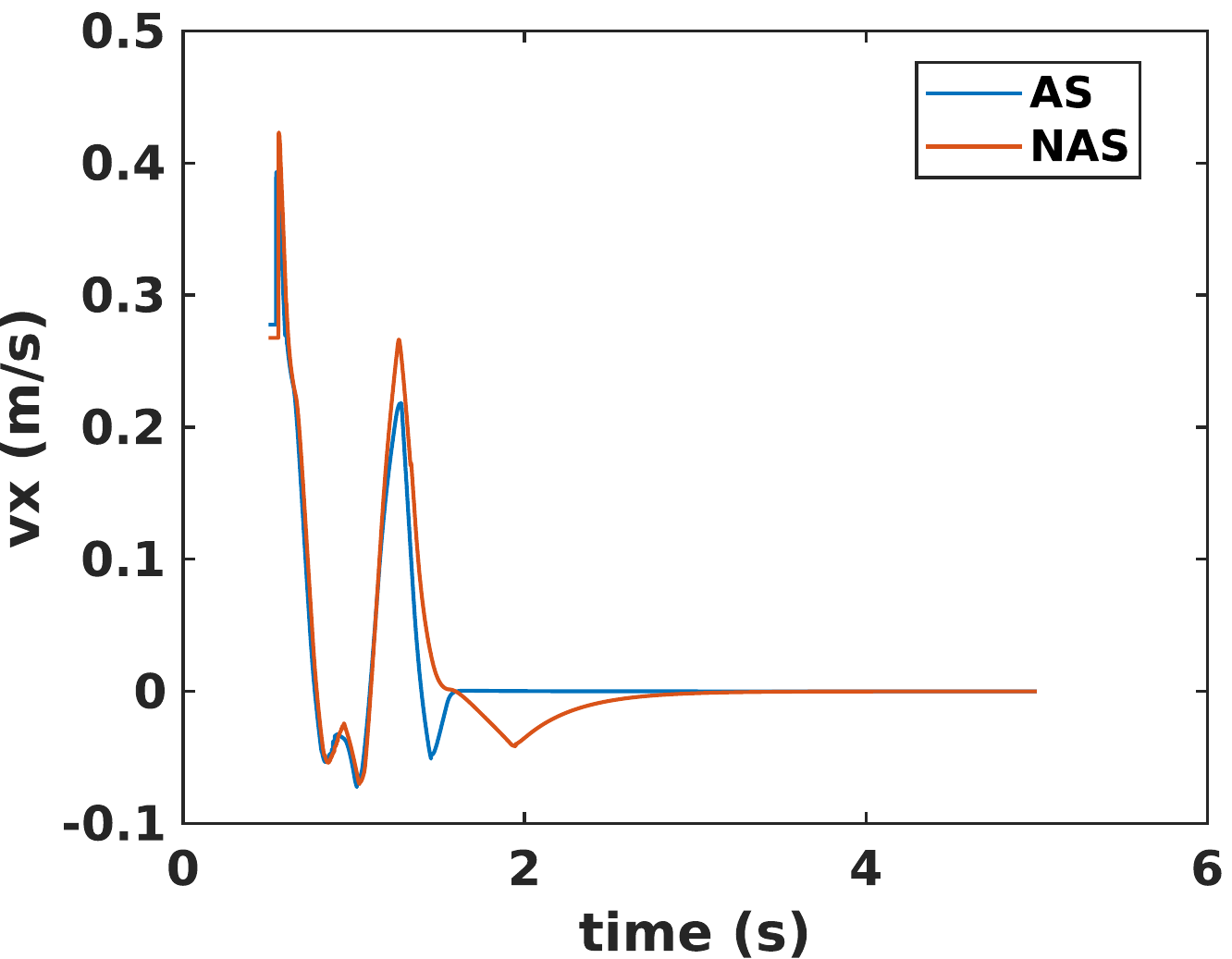}
    \caption{Fig. 8. vx in landing phase.}
    \label{vx in landing phase}
  \end{minipage}%
  \begin{minipage}[t]{0.5\linewidth}
    \centering
    \includegraphics[scale=0.29]{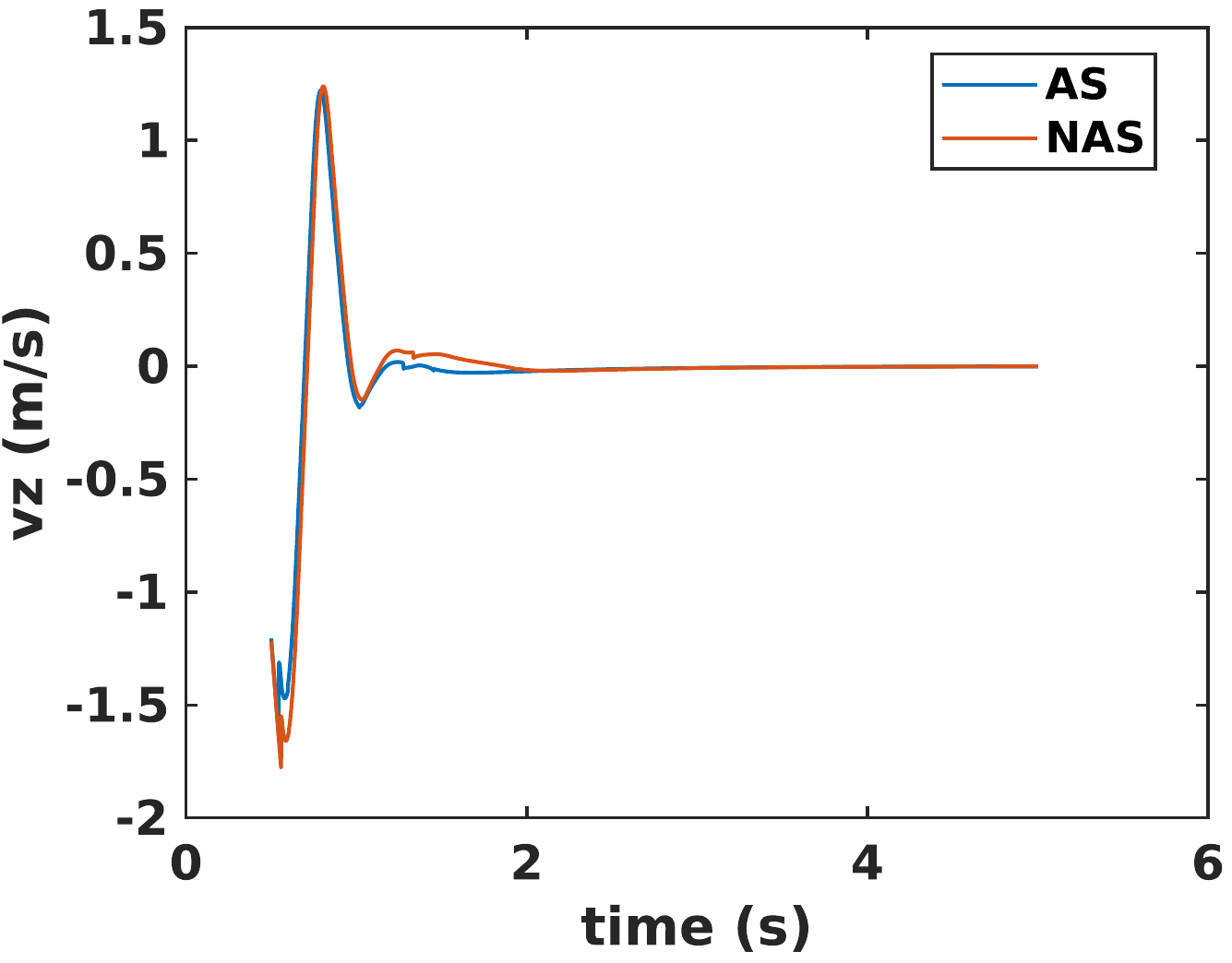}
    \caption{Fig. 9. vz in landing phase.}
    \label{vz in landing phase}
  \end{minipage}
\end{figure}

\begin{figure}[htbp]
  \begin{minipage}[t]{0.5\linewidth}
    \centering
    \includegraphics[scale=0.29]{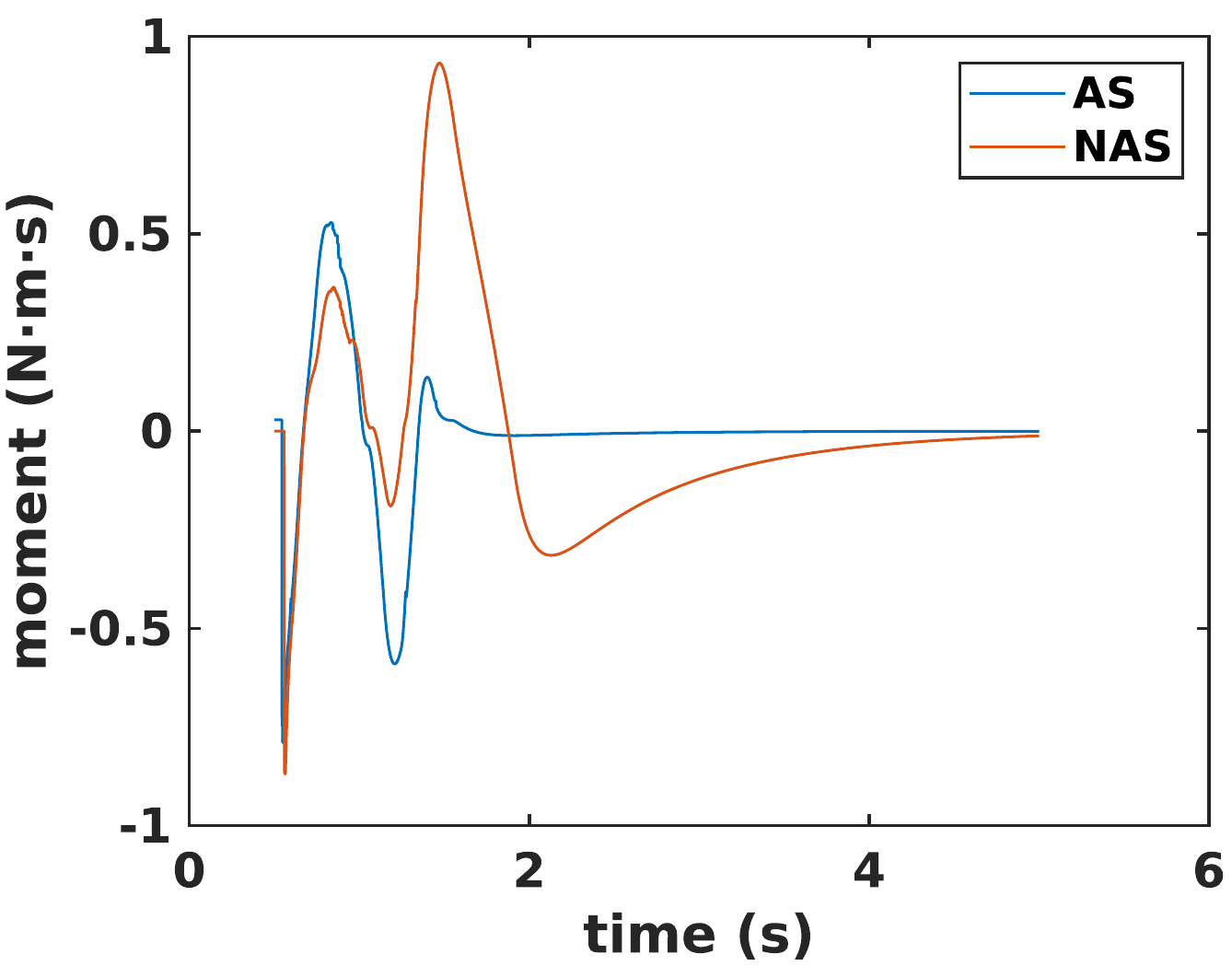}
    \caption{Fig. 10. Pitch moment.}
    \label{Pitch moment}
  \end{minipage}%
  \begin{minipage}[t]{0.5\linewidth}
    \centering
    \includegraphics[scale=0.29]{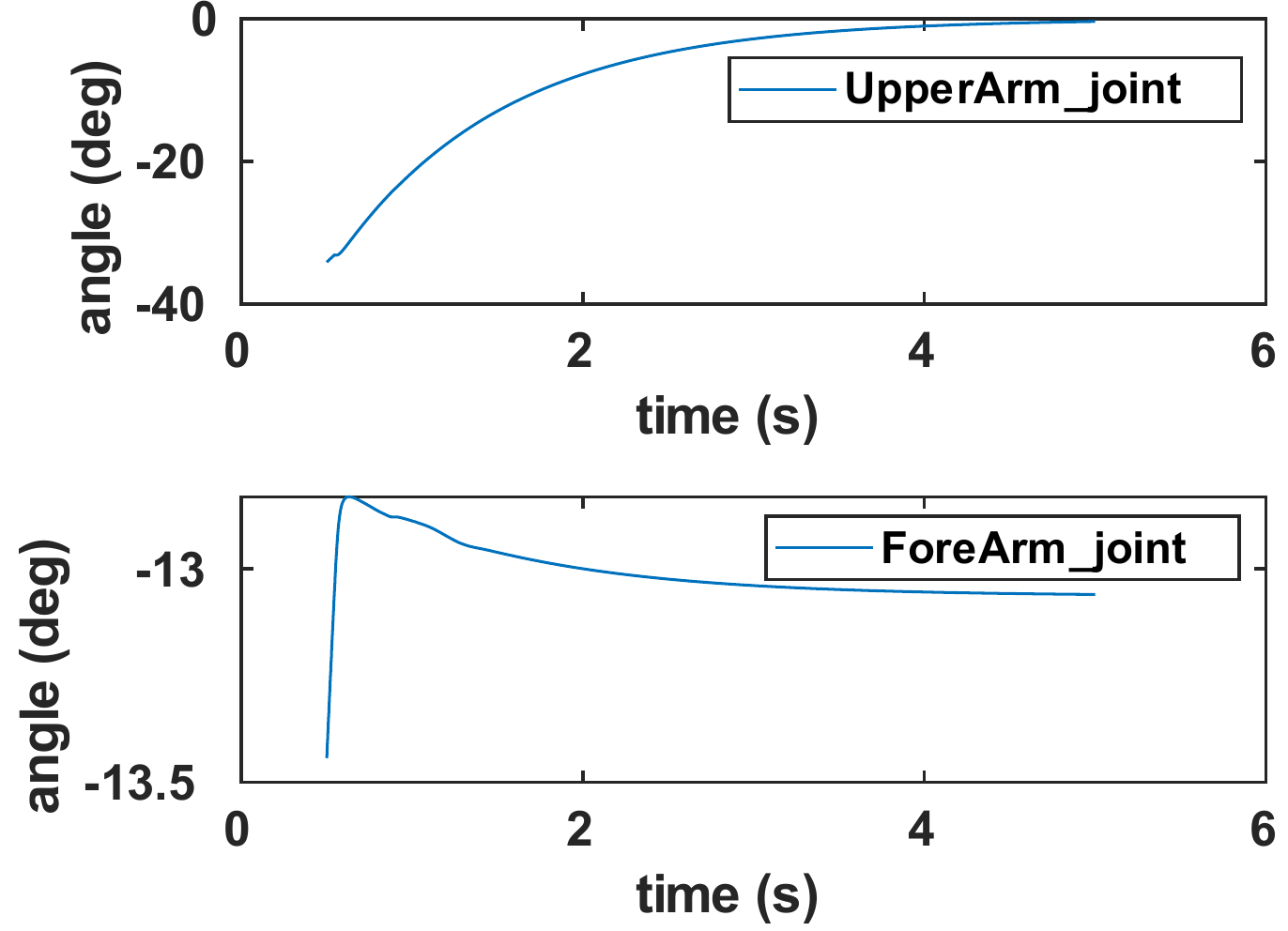}
    \caption{Fig. 11. Arm angles in landing phase.}
    \label{Arms angle in landing phase}
  \end{minipage}
\end{figure}

We acquired the COM velocity of the landing phase (Fig. \ref{vx in landing phase}, Fig. \ref{vz in landing phase}), pitch moment of COM (Fig. \ref{Pitch moment}) and joint angles of arms (Fig. \ref{Arms angle in landing phase}), and then we calculated the TLSEs of vx, vz and pitch moment of both AS and NAS(TABLE \ref{TLSE}) and the stability time of both AS and NAS(TABLE \ref{Stabilization time}).

We observed that the frequency and amplitude of the oscillation of vx, vz and pitch moment is more severe than that of the AS from Fig. \ref{vx in landing phase} \ref{vz in landing phase} \ref{Pitch moment} and TABLE \ref{TLSE}. It also can be seen that stabilisation time of AS is reduced by a factor of two compared to that of NAS from TABLE \ref{Stabilization time}. As shown in Fig. \ref{Arms angle in landing phase}, the arms have been swinging until the robot reaches static during the whole landing phase, which demonstrates the importance of the swing arms in restoring balance\cite{pijnappels2010armed}.

\begin{table}[htbp]
	\caption{TABLE II. TLSE}
	\label{TLSE}
	\begin{center}
		\begin{tabular}{|c|c|c|}
			\hline
			Quantity & AS & NAS\\ \hline
			vx & 113 m/s & 145 m/s\\ \hline
			vz & 449.3 m/s & 505.3 m/s \\ \hline
			pitch moment   & 300 N·m·s  & 797 N·m·s\\ \hline
		\end{tabular}
	\end{center}
\end{table}

\begin{table}[htbp]
	\caption{TABLE III. Stabilization time}
	\label{Stabilization time}
	\begin{center}
		\begin{tabular}{|c|c|}
			\hline
			AS & 1.582s\\ \hline
			NAS & 4.526s\\ \hline
		\end{tabular}
	\end{center}
\end{table}

\subsection{Energy consumption results}
\begin{figure}[htbp]
	\begin{center}
		\includegraphics[width=3.0in,trim=10 0 0 0]{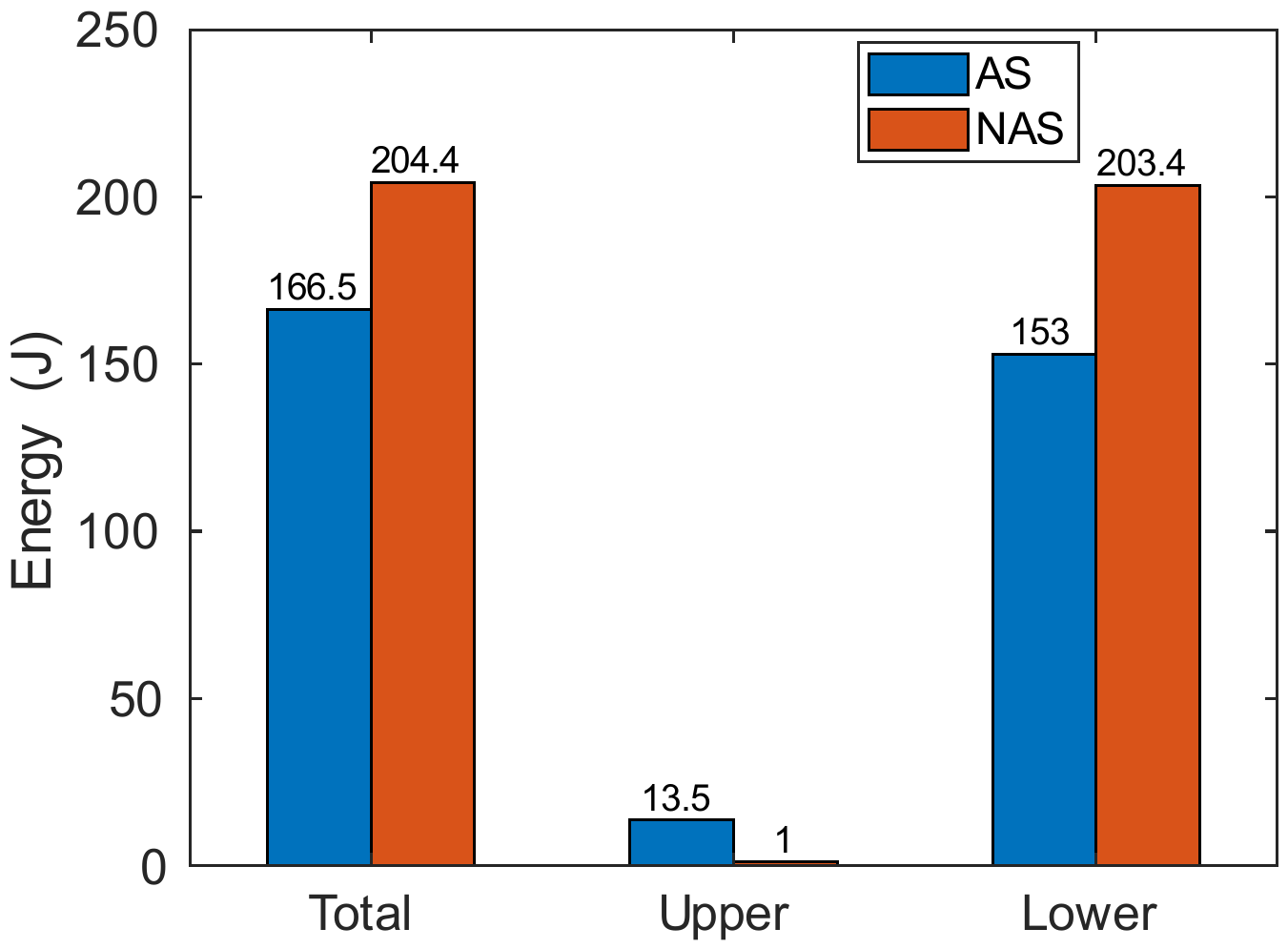}
	\end{center}
	\caption{Fig. 12. Energy cost.}\label{Energy cost}
\end{figure}

In this section, we defined the energy consumption as sum of joint work and obtain the total energy consumption of both AS and NAS(Fig. \ref{Energy cost}). We observed that the total energy consumption of the AS is reduced by more than 20 $\%$ to that of the NAS. Moreover the energy saved (50.4J) of the lower extremities (legs and trunk) is much greater than the energy consumed (12.5J) of the upper extremity (arms) in the AS, which indicates that the total energy saved is primarily due to the swing arms.




\section{CONCLUSION AND FUTURE WORK}
This paper presented the jumping motions of a bipedal robot in both cases AS and NAS and analysed the results of the jumping motions from three aspects of agility, stability, energy consumption and it was concluded that all three of these aspects can be improved by introducing swing arms.

Despite the expected results of the experiment, the theoretical basis of the relationship between swing arms and the motion performance of bipedal robot remains unclear, moreover we will verify these results on the physical robot Purple V1.0. In the future, we will explore this theoretical basis and obtain better arm-swing trajectory to maximize the role of the swing arms. In addition, we will apply the swing arms to various more complex motion such as running and continuous jumping.

\addtolength{\textheight}{-12cm}   



\bibliographystyle{IEEEtran}
\bibliography{ref}

\end{document}